\documentclass[twocolumn]{article}
\usepackage[authoryear]{natbib}
\usepackage{times}
\usepackage{helvet}
\usepackage{url}
\usepackage{courier}
\usepackage{multirow}
\usepackage{graphicx}
\usepackage{tabularx}
\usepackage{authblk}
\usepackage[htt]{hyphenat}

\title{Automatic Classification of Object Code Using Machine Learning}
\author[1,2]{John Clemens}
\affil[1]{University of Maryland, Baltimore County (UMBC), Baltimore, Maryland \texttt{$<$clemej1@umbc.edu$>$}}
\affil[2]{Johns Hopkins University Applied Physics Laboratory (JHU/APL), Laurel, Maryland \texttt{$<$john.clemens@jhuapl.edu$>$} }
\date{}

\begin{document}
\maketitle
\begin{abstract}
Recent research has repeatedly shown that machine learning techniques can 
be applied to either whole files or file fragments to classify them for
analysis. We build upon these techniques to show that for samples of 
un-labeled compiled computer object code, one can apply the same type of 
analysis to classify important aspects of the code, such as its target
architecture and endianess. We show that using simple byte-value histograms
we retain enough information about the opcodes within a sample to 
classify the target architecture with high accuracy, and then discuss 
heuristic-based features that exploit information within the operands 
to determine endianess.  We introduce a dataset with over 16000 code 
samples from 20 architectures and experimentally show that by using our 
features, classifiers can achieve very high accuracy with relatively small 
sample sizes.
\end{abstract}

\section{Motivation}

Digital forensics remains largely a manual process requiring detailed 
and time consuming analysis by experts within the field.  In particular, 
the analysis of computer executables, either for forensic analysis, reverse 
engineering, or malware detection, remains a time consuming task
as the level or expertise needed to understand compiled object code is quite 
high.  Additionally, the explosion of different types of devices (cell phones, 
complex routers, smart sensors, the internet of things (IoT)) means that
experts are no longer dealing with just one computing architecture, but instead
are seeing a myriad of executable code (firmware, mobile apps, etc.) traversing
their networks and showing up in forensic and malware samples.  Even generic
desktop workstations contain object code for architectures other than the 
main CPU. These can include GPU-enabled programs, firmware for network cards and
other devices which contain embedded CPUs (\cite{blanco2012}, \cite{nicreverse}), 
management co-processors (\cite{battfirm}), and USB drivers for devices that contain
their own processors for services like data compression or encryption.  The 
object code for these devices is often stored in files with non-standard headers
or embedded inside driver object files. Analysts are seeking tools to
jump-start the analysis process by automatically labeling unknown samples. 

Plenty of recent research has shown that raw byte frequency analysis can be 
used to classify files and file fragments. These analyses fall short in 
two areas when applied to object code.  First, by taking the entire sample
into consideration, they include file meta-data into their analysis.
In many cases this is beneficial, but there are a few cases where this might
be a concern.  For example, the sample itself may be incomplete (a partial 
forensic disk recovery or a partial packet capture), not trustworthy 
(deliberate obfuscation by malware), or simply have no meta-data (firmware, 
reverse engineering, raw instruction traces from virtual machines).  Ideally 
analysts want a classifier that relies solely on the object code itself, 
ignoring any
meta-data that may (or may not) be present.  Secondly, the analysis from 
most previous work stops one level above what we believe is possible.  
These systems will identify a sample as containing object code, but 
won't give any more information than a general file label. When possible, 
we should label the sample with information about the type of object code 
the sample contains. 

We propose methods that apply machine learning techniques to automatically 
classify an object code sample with its target architecture and endianess.  
Such a system automates the first phase of object code analysis, allowing the 
analyst to jump directly to decoding the instructions and determining 
intent.

The rest of this paper is structured as follows: The next sub-section 
discusses related research. In the Hypothesis section we attempt to 
formalize the problem of architecture and endianess classification. Next we 
discusses the intuition behind our proposed solutions, and then go over our
experimental design and results. We conclude with a discussion of the results and potential
follow-on work.

\subsection{Related Research}
\label{sec:related}
Many systems exist to determine the type of binary code a file may contain. 
The simplest systems rely solely on the file name or file extension. However, 
most systems rely on the contents of a ``file header'' at a known location within
the file (normally at the beginning) which includes metadata about what type of
file it is, such as a document, picture, or executable. The UNIX {\em file} 
command uses a database of ``magic'' values 
at known offsets within the file to classify the file type.  In the case of 
executables or other object code, these file type (ELF, PE, etc.) headers 
contain fields with information such as the target architecture, word size, 
and endianess. Each of these systems uses some form of meta-data 
(file header, signature, or filename) that may not be available to an analyst.

\cite{mcdaniel2003content} were among the first to
propose using characteristics derived from the contents of an entire file to do 
classification.  They used byte-value histograms as one of their representations
and performed statistical analysis to classify files.  This inspired many more 
researchers to use other methods including n-gram analysis and SVMs to tackle
the same problem. Examples include \cite{fitzgerald2012}, 
\cite{li2010}, \cite{fileprints}, and \cite{xiebyte}.  \cite{sceadan} produced the Sceadan tool 
which builds upon much of this earlier work.  This line of 
research has concentrated on differentiating diverse file types from each
other.

Relating specifically to architecture classification, \cite{chernov} attempt 
to automate the analysis of custom virtual machines
used by malware.  Their system uses opcode frequency counts 
as part of their analysis system to help defeat code obfuscation within the 
custom virtual machine.  Similarly, \cite{rad2012} show 
that opcode frequency code counts can be used to find mutated forms of the
same malware.  They rely on knowledge of the underlying physical 
system's opcodes as an indicator of program similarity. 

\cite{sickendick} describes a system for firmware disassembly
including file carving and architecture detection using machine learning. 
For architecture detection, he adapts the method \cite{kolter}, 
used for malware detection. The information gain for each byte value 4-gram 
in the training set is calculated, and the top 500 4-grams are used as a 
feature vector for a DecisionTree and an SVM classifier.  This work is limited 
to four architectures common to SCADA devices and makes no attempt to classify
different endianess with the same architecture.

Binwalk (\cite{binwalk}) is a popular firmware 
analysis tool that includes two techniques to identify object code.  
When run with the '-A' option, Binwalk looks for architecture specific 
signatures indicative of object code.  Currently, Binwalk's architecture 
signature detection includes 33 signatures from 9 different 
architectures. However, Binwalk simply reports every place it finds a 
signature and leaves it up to the user to make a classification decision based 
upon that information.  Binwalk also includes a '-Y' option which will attempt 
to disassemble code fragments using the Capstone (\cite{capstone}) disassembly
framework configured for multiple architectures.  Binwalk currently 
supports 9 configurations of 4 unique architectures for disassembly.  Notably, 
both methods can potentially indicate endianess as well as architecture.

Binwalk's methods are effective in a wide variety of use cases, but are
not without their limitations.  Signature based methods can lead to false
positives if the byte signatures are not unique when compared to other architectures.  
Evidence of such collisions exists in the Binwalk code itself, where 
a comment mentions that some 16-bit MIPS code signatures are often detected 
in ARM Thumb code.  Disassembly of a fragment can also cause issues.  There 
is at least one case (i386 versus x86\_64) where both architectures could 
disassemble the same fragment of code without error. Both techniques rely
on previous knowledge of the architecture, and in the case of active
disassembly, complete knowledge and support in a disassembler framework. 
The technique presented in this paper takes a more holistic approach, and
is able to classify architectures, both virtual and physical, for which there 
are samples, even if information about the architecture is incomplete.

\section{Problem}
\label{problem}
We aim to automatically classify two characteristics of computer object code:
\begin{itemize}
	\item{{\em Architecture}: The unique encoding of the computer's 
				instructions.}
	\item{{\em Endianess}: The way the code expects multi-byte data to
				be ordered when in memory.}
\end{itemize}

Computer object code consists of a stream of machine instructions encoded
as a string of bytes.  The instruction stream is loaded into memory and stored
in the native endianess of the processor.  The processor fetches instructions
from the instruction stream in memory, and then decodes and executes them.  
Computers share the same {\em architecture} if they use
the same (or similar) encodings for these machine instructions.  The
encoding of the instructions is referred to as an {\em instruction set}.  Some
architectures define fixed-length instruction encodings while others define
variable-length instruction encodings. This makes it impossible to determine
the boundaries of instructions within an instruction stream without knowing the 
target architecture.

Machine instructions consist of two parts: the {\bf opcode} specifies which 
instruction the processor is to execute, and {\bf operands} which specify 
what data (or pointers to data) that the instruction applies to.  Opcodes are 
the byte representation of the instruction and are specified by the 
architecture.  Operands can be many things including encoded register values, 
memory locations, and direct data values.  While opcode encodings are unique 
to a specific architecture, operands vary with the data and flow of the 
particular program. To accurately classify the architecture, one should 
isolate its opcodes.

Endianess refers to the way the architecture stores multi-byte data in memory.  
There are two ways multi-byte values may be encoded: least 
significant byte first (little endian) or most significant byte first (big 
endian).\footnote{There is also ``mixed endian'', but that is no longer in wide 
use and not considered for this analysis.} Most 
architectures define an endianess, so knowing the architecture automatically 
infers the endianess.  However, some architectures (e.g. MIPS, ARM, Power) 
can be configured to use either endianess at runtime, and thus a proper 
classification must also determine the endianess of a sample for those 
architectures. 

Since endianess deals with the layout of data in memory, it is difficult to 
determine from a sample of object code alone. However, operands may contain
immediate values and/or address values which are encoded in the native endianess
of the architecture when stored in memory or on disk. Any system that classifies
endianess from an instruction stream may be able to extract that 
information from the portion of the object code used for operands. 

\section{Hypothesis}
\label{solution}
Previous research (\cite{mcdaniel2003content}) has shown that byte-value 
histograms over an entire file can be useful when classifying a file's type. 
We propose to apply this same basic technique to the object code 
embedded within a sample.  We deliberately ignore the rest of the file as it 
may contain meta-data that is either not present or not trustworthy within a 
given scenario. 

Examples from some known architecture encodings gives us reason to believe that
a byte-value histogram will be useful for classification.
The `amd64' architecture is a 64-bit extension of the `i386' architecture, and 
uses a special ``prefix'' byte for every instruction that uses 64-bit operands.
This byte has the high 4-bit nibble set to b`0100' and the lower four bits 
change depending on the rest of the instruction.  One would expect a 
byte-value histogram for a sample from the amd64 architecture to contain 
many values that start with `0x4'.  ARM instruction encoding specifies the upper 4 bits of each
instruction start with `condition codes'.  For most instructions, these are
set to b`1110', which means `always execute'.  Therefore, one would expect
that a byte-value histogram for ARM systems to contain many values that 
start with `0xE'.  Intuitively, a machine learning algorithm should be able
to accurately classify between these two architectures based solely on a 
byte-value histogram.

More generally, in order for a byte-value histogram to be useful for 
classifying object code, the uniqueness of the architecture's opcodes must be 
preserved within the histogram.  To demonstrate this is possible, we need an 
estimation of
how likely an opcode is to influence each byte within the code section. We 
call this the {\em opcode density} of the architecture, and it is 
calculated by the formula:

$$ Opcode\_Density = \frac{length\_of\_opcode}{average\_instruction\_length} $$

For fixed-length instruction set architectures, the instruction length is fixed
(normally 32 or 64 bits depending on the architecture's word size), and the 
opcode takes up between 6 and 12 bits, depending on the instruction.  To use
MIPS as an example, the instruction length is 4 bytes, and the opcode is 6 bits
long, for an opcode density of approximately 19\%.  Practically, this means the first byte 
of every instruction (one in four bytes) will have the opcode encoded in its 
top 6 bits, heavily influencing its value.
Similar analysis can be carried out using the SPARC and Alpha architectures, 
where the opcode is encoded in 8 bits, and ARM (8-bit opcodes + 
4-bit condition codes).  Even if we assume that the operands in the object code
are random values, one can see that for fixed length instruction encodings
one in four byte-values within the object code will be heavily influenced
by the opcode value.

For variable length instruction sets the analysis is more difficult, as we no
longer know the ratio of opcodes to total instruction length.  Intel i386 
opcodes have a minimum length of one byte (but can be two or more). \cite{blem2013} 
show that on average, the i386 architecture for general desktop 
workloads has an instruction length of 3.4 bytes.  This means that even if we 
assume one-byte opcodes, our opcode density is approximately 30\%, or at the 
very least it is higher than most fixed-length instruction encodings for a 
typical workload.

These rough calculations give us some confidence that a byte-value histogram
can preserve information about the opcode encoding, and thus can be used for 
architecture classification.

\subsection{Endianess}

Unfortunately, determining endianess is impossible with a byte-value histogram
alone. Determining endianess requires byte adjacency information, and adjacency
information is lost in the conversion to the histogram. Therefore, in order
to determine endianess, we need another set of features that can preserve 
byte ordering information.

One approach would be to generate a 2-byte-value (bi-gram) histogram.
While this may encode adjacency information, it would explode our feature 
space from 256 dimensions to 65536, adding a large amount of computational
complexity. Also, despite the intuition, our experiments show that this 
approach is not useful for determining endianess.

In the previous analysis we treated the operands for a sample as random 
noise.  While convenient for that analysis, at least some instructions 
encode `immediate' data within their operands.  These operands are stored
in the object code in native-endian format.  We aim to exploit this
information to determine endianess using a small set of heuristics. 

On machines without an increment instruction, one common operation when 
incrementing by a small value is to use an add instruction with an immediate 
operand of 1.  On big endian machines, one is encoded in 32 bit as 0x00000001, 
while on little endian machines it is encoded as 0x01000000.  This provides us 
with a heuristic: if we scan the object code for the 2-byte strings `0x0100' 
and `0x0001', then the latter should occur more often in little endian samples
and the former should occur more often in big endian samples.  This could
be repeated for other small values.  Another common immediate value encoded in
operands are addresses.  Some addresses, typically for stack values, are high
up in the address space and start with values like 0xfffe.  
Again, these addresses are stored differently on big endian versus little 
endian machines, and a scan for both values 0xfffe and 0xfeff can be used as 
another indicator of endianess.

\begin{figure}
	\centering
	\includegraphics[scale=0.45]{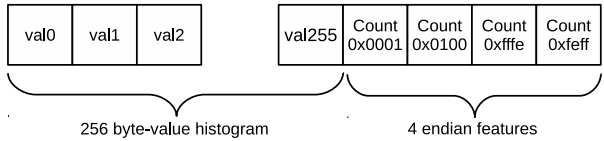}
	\caption{Layout of the full 260-dimension feature vector.}
	\label{figure:featurevec}
\end{figure}

We propose to use these four heuristically derived 2-byte frequency counts 
(`0xfffe',`0xfeff',`0x0001',`0x0100') as four new ``endian'' features to augment 
the byte-value histogram, as shown in Figure \ref{figure:featurevec}. We 
demonstrate that these features add the ability to predict endianess with 
minimal computational overhead.

\section{Experiments}
\label{experiments}
We tested the theory that our features are sufficient to classify architecture 
and endianess by creating a dataset of sample object code, 
generating the representative feature vectors, and then training machine 
learning models using our features. 

\subsection{Dataset}

The Linux operating system has been ported to many different architectures
since its inception, and provides a rich starting point for our dataset.
A typical distribution installs anywhere from 600 to 1300 files that contain 
compiled object code for the supported architectures.  A large number of our 
samples come from the Debian Linux distribution for different architectures. 
To augment the dataset beyond what is available within Linux systems, 
we collected samples of Arduino code that targets the AVR line of 8-bit 
micro-controllers as well as CUDA samples that target the nVidia line of GPUs.
All sample files in this data set are ELF files, and object code identified 
by using the PyBDF (\cite{pybfd}) library to parse ELF section information.

A summary of the resulting dataset with samples from 20 different architectures
is shown in Table ~\ref{table:dataset}.  Of particular interest to endianess 
classification is the inclusion of `mips' and `mipsel' as two different
classes.  As both classes use the exact same opcodes, the only difference 
between the samples is the endianess of values within their operands. 

\begin{table}[t]
	\centering
	\begin{tabular}{ l || r  r  r }
		\hline
		Architecture & \# Samples & Wordsize & Endianess \\
		\hline
		alpha & 1,383 & 64-bit & Big \\
		hppa & 625 & 32-bit & Big \\
		m68k & 1,296 & 32-bit & Big \\
		arm64 & 1,134 & 64-bit & Little \\
		ppc64 & 823 & 64-bit & Big \\
		sh4 & 822 & 32-bit & Little \\
		sparc64 & 752 & 64-bit & Big \\
		amd64 & 965 & 64-bit & Little \\
		armel & 960 & 32-bit & Little \\
		armhf & 960 & 32-bit & Little \\
		i386 & 967 & 32-bit & Little \\
		ia64 & 650 & 64-bit & Little \\
		mips & 960 & 32-bit & Big \\
		mipsel & 960 & 32-bit & Little \\
		powerpc & 992 & 32-bit & Big \\
		s390 & 649 & 32-bit & Big \\
		s390x & 653 & 64-bit & Big \\
		sparc & 648 & 32-bit & Big \\
		cuda & 17 & 32-bit & Little \\
		avr & 596 & 8-bit & Little \\
		\hline
		Total & 16,785 & & \\
		\hline
	\end{tabular}
	\caption{Dataset statistics for all 20 architectures. Note that these 
		reflect the samples that are in the dataset, not the full 
		capabilities of the architecture.  For example, there can be 
		HPPA systems that are 64-bit, and ARM, MIPS, and PowerPC can 
		all be configured as either little endian or big endian.}
	\label{table:dataset}
\end{table}

As with all datasets, this one could be improved.  All samples except the CUDA
samples are compiled with GCC.  A different compiler might use a different
mix of opcodes and thus have a different signature.  Additionally, there 
are many more 8 and 16-bit architectures than what are represented here. We 
hope to augment this dataset over time to add more diversity among the samples.

\subsection{Feature Generation}

As described above, we will use a feature vector that
contains a byte-value histogram of the code section augmented with four 
additional
counts of specific values we will look for to indicate endianess.  The 
layout of the feature vector is shown in Figure ~\ref{figure:featurevec}.

When preparing the samples, we can choose to have one feature vector per sample
file, or we can choose to extract the code from each file into one big pool  
and draw equal-sized samples from the global pool.  The latter approach might 
be beneficial to avoid an issue where an individual file's code sections are 
tiny, and thus has mostly zero values in its histogram.  However, the approach of
one-sample-per-file is a more realistic scenario in the field.
For this paper, one feature vector is generated per sample file. 

The byte-value histogram is generated by scanning every sample file for 
all sections labeled as executable code, and then reading those sections
one byte at a time to generate our byte-value histogram.  When the
entire file has been processed, the histogram values are normalized by dividing
each value by the number of bytes of code within that file.  These make up the 
first 256 entries in the feature vector. The four additional endianess values
are calculated by a linear scan of each code section for the specific 
two-byte values.  These counts are normalized over the size of the code 
sections within the file as well. All parts of the file that do not contain
object code, as defined by the ELF section's \emph{CODE} flag (or, in the case 
of CUDA code, an ELF section named \emph{.nv\_fatbin}), are explicitly 
excluded from the feature vectors.

In addition to generating samples that use the entire code section within the 
sample file, we also want to test against object code fragments of varying size.
To generate those feature vectors, the same procedure
is followed except that the byte values are taken as a random sampling of the
code bytes up to the desired size (or the end of the code 
section). Random sampling removes any bias that may present itself by 
continuously using the beginning of each code section. For these feature 
vectors, the endian feature counts are also generated using random 2-byte 
sampling of N offsets within the code section, where N is the maximum size of 
the sample.  The appropriate feature count is incremented if the random 2-byte 
sample matches one of the specific 2-byte values we're searching for. These 
counts are also normalized to the number of code bytes used within the sample.

To test the effectiveness of 2-byte bi-grams, we generate 64k-entry feature 
vectors for the `mips', and `mipsel' classes.  We can then compare the 
results when using this data subset to the overall results using our four
endian features. 


\section{Results}
\label{results}

\begin{table*}[p]
	\centering
	\begin{tabular}{ l  c  l || c  c }
		\hline
		Trained Model & Multi-class Strategy & WEKA Name & Histogram & Hist + Endian \\
		\hline
		1-NN & Inherent & IBk & 89.3238\% & 92.7256\% \\
		3-NN & Inherent & IBk & 89.8660\% & 94.9002\% \\
		Decision Tree & Inherent & J48 & 93.2976\% & 98.0697\% \\
		Random Tree & Inherent & RandomTree & 87.8046\% & 92.9461\% \\
		Random Forest & Inherent & RandomForest & 90.4617\% & 96.4373\% \\
		Naive Bayes & Inherent & NaiveBayes & 92.5827\% & 95.8951\% \\
		BayesNet & Inherent & BayesNet & 89.5144\% & 92.2252\% \\
		SVM (SMO) & 1-vs-1 & SMO & 92.7256\% & 98.3497\% \\
		Logistic Regression & Inherent & SimpleLogistic & 93.0831\% & 97.9386\% \\
		Neural Net & Inherent & MultilayerPerceptron & 94.0244\% & 97.9565\% \\
		\hline
	\end{tabular}
	\caption{10-fold stratified cross validation accuracy for various models
                 using the byte-value histogram alone, and the byte-value 
		 histogram augmented with heuristic-based endianess attributes.}
	\label{table:accuracy}
\end{table*}

We used the generated feature vectors to train a set of common multi-class 
classifiers available in WEKA (\cite{weka}). The models chosen are inherently
multi-class, with the exception of the SVM (SMO) model which uses a 
series of 1-versus-1 comparisons to choose the final class. The results are 
summarized in Table ~\ref{table:accuracy}  
which shows the 10-fold stratified cross validation accuracy for the chosen 
classifiers.  Of note, the linear-based classifiers (Logistic Regression, SVM) 
and the Decision Tree seem to have the greatest accuracy, but all classifiers do
very well.  This clearly shows that there is enough unique information about 
the architecture exposed within the byte histogram to accurately classify 
object code in nearly all instances.
 
Table ~\ref{table:fscore} shows the F-Measure values broken down by class for
the Logistic Regression classifier. F-Measure is the harmonic mean of Precision
and Recall.  Higher F-Measure values indicate better classification performance,
and a value of 1.0 would be perfect classification. The chart shows that the 
majority of the classification errors are caused in the `mips' and `mipsel' 
classes when we do not include our four endianess features and rely solely on 
the byte histogram. The dramatic improvement in F-Measure with these features 
shows that they are indeed useful heuristics for determining endianess. Note
that CUDA F-Measure scores suffer from the small number of CUDA samples 
available within the dataset.

\begin{table}[t]
	\centering
	\begin{tabular}{ l || r  r }
	\multicolumn{1}{l}{} & \multicolumn{2}{c}{F-Measure} \\
	\hline
	Architecture & Histogram & Hist + Endian \\
	\hline
alpha	& 0.992	& 0.997 \\
hppa	& 0.994	& 0.993 \\
m68k	& 0.995	& 0.993 \\
arm64	& 0.987	& 0.994 \\
ppc64	& 0.995	& 0.996 \\
sh4	& 0.993	& 0.993 \\
sparc64	& 0.987	& 0.993 \\
amd64	& 0.987	& 0.990 \\
armel	& 0.998	& 0.998 \\
armhf	& 0.994	& 0.996 \\
i386	& 0.995	& 0.998 \\
ia64	& 0.995	& 0.995 \\
mips	& 0.472	& 0.884 \\
mipsel	& 0.476	& 0.886 \\
powerpc	& 0.990 & 0.989 \\
s390	& 0.998	& 0.998 \\
s390x	& 0.998	& 0.998 \\
sparc	& 0.988	& 0.988 \\
cuda	& 0.444	& 0.516 \\
avr	& 0.926	& 0.936 \\
	\hline
	\end{tabular}
	\caption{Resulting per-class F-Measure for the Logistic Regression 
		model.  Note the increase in score for the mips and mipsel 
		targets with the addition of endian features. Other models 
		show a similar pattern.}
	\label{table:fscore}
\end{table}

These classifiers are mostly trained with their default parameters.
One notable exception to this is the Neural Network classifier,
which suffers from overfitting when adding the endian features with the
default network structure of 260x140x20.  A partial grid search over the 
number of epochs and the number of hidden nodes suggest a network 
configuration of 260x66x20 with 100 epochs results in performance 
in line with the other classifiers.  See Table ~\ref{table:params} for the 
full breakdown of all parameters used to generate these results. Parameters 
for each classifier could undoubtedly be tuned further for even greater 
classification performance.  

Finally, Table ~\ref{table:bigrams} shows the F-Measure of two models 
classifying `mips' versus `mipsel' using a 64k bi-gram histogram versus our
260 feature byte histogram and endian features.  Surprisingly, the bi-gram 
encoding appears to preserve much less endian information than our simpler
heuristic-based method despite the much higher computational overhead of its
larger feature vector. 

\begin{table}[t]
	\centering
	\begin{tabular}{ l || c  c || c  c }
	\hline
	\multirow{2}{*}{Trained Model} & \multicolumn{2}{c||}{64k Bi-grams} &  \multicolumn{2}{c}{Hist + Endian} \\
	& mips & mipsel & mips & mipsel \\
	\hline
	Random Forest & 0.530 & 0.453 & 0.721 & 0.681 \\
	Decision Tree & 0.477 & 0.476 & 0.897 & 0.897 \\
	\hline
	\end{tabular}
	\caption{Comparison of the F-Measure results when using straight 
		bi-grams over the four heuristic endianess features proposed in 
		this paper.  The performance of the classifier when using the 
		proposed features is significantly better while being less 
		computationally expensive.}
	\label{table:bigrams}
\end{table}
\subsection{Sample Size}

\begin{figure*}[p]
	\centering
	\includegraphics[scale=0.70]{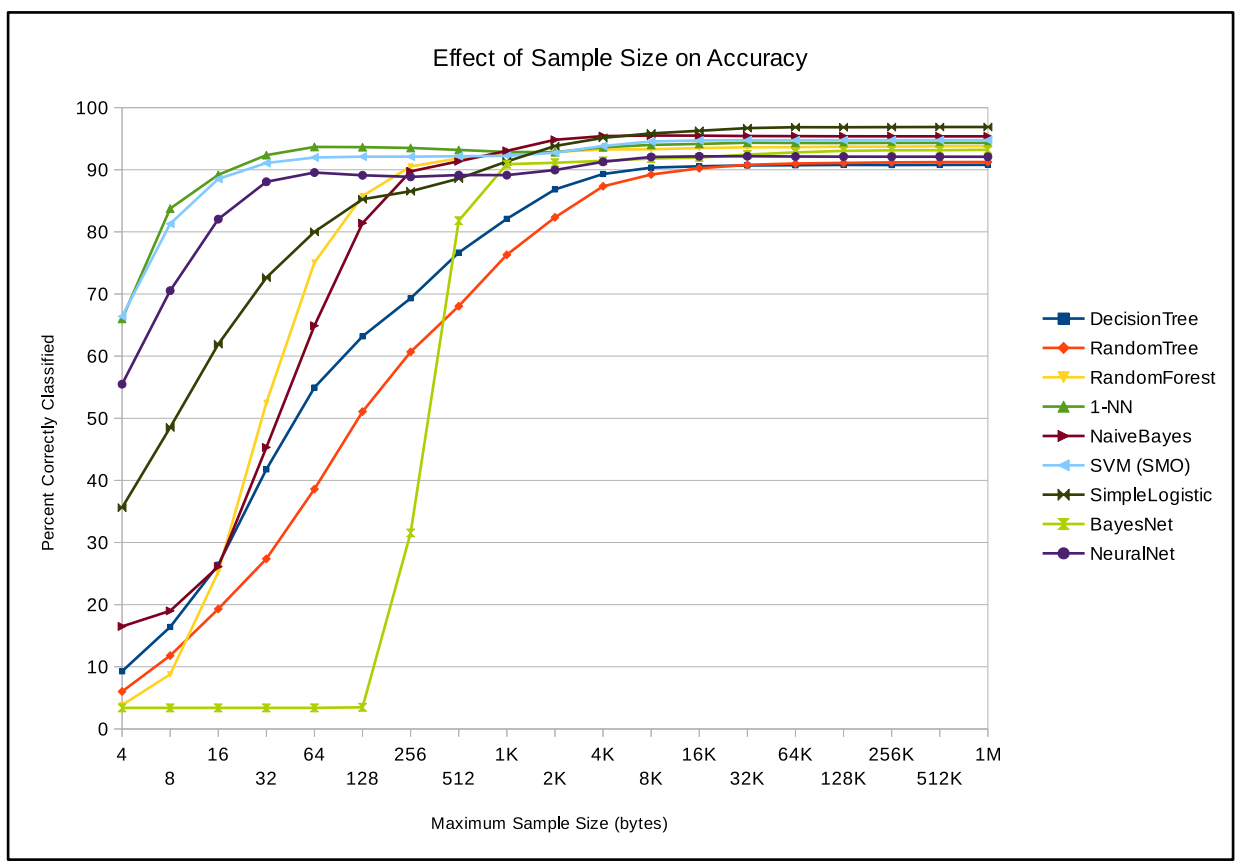}
	\caption{10-fold cross validation accuracy of the classifiers for 
		different maximum sample sizes.  Note that for both SVM and 
		1-NN, the accuracy approaches 90\% with only 16 bytes of 
		sample data.  By 8KB, all classifiers are near or above 90\% 
		accuracy.}
	\label{figure:sample-size}
\end{figure*}

The above results achieve high accuracy using the every byte of object code
available within each sample.  Another question is how large of a 
sample fragment do you need to achieve high accuracy. This is a useful
metric for analysts who often deal with incomplete fragments of samples.  
To test this, we generate new feature vectors from our samples using 
maximum sample sizes of four bytes up to one megabyte using the random sample 
methodology explained earlier.  We then ran each of these size-based feature 
sets through the models trained on the full-sample instances.  The results are 
summarized in Figure ~\ref{figure:sample-size}. These results show that for 
both the SVM and 1-NN classifiers, one can achieve very high accuracy even for
tiny amounts of sample data, and that by 8KB, nearly all classifiers are
above 90\% accuracy.

\section{Discussion and Further Work}
\label{discussion}
We have shown that machine learning can be an effective tool to classify the 
target architecture of object code. As this method is independent of 
potentially misleading meta-data, it provides both a way to verify existing 
meta-data and a way forward when no meta-data is present. We have developed 
heuristics that can be used to predict the endianess of code.  Of the 
classifiers tested, SVM and nearest neighbor approaches appear to provide good
classification performance regardless of fragment size. 

Going forward, we would like to expand our current architecture dataset to
include a more varied sampling of architectures.  We intend to include more 
embedded platforms, microcontroller code, and more GPU samples.  We will also
include samples using different compilers than GCC, including LLVM/Clang and 
Microsoft Visual Studio, to make sure that different code generation engines 
do not effect the overall classification performance.

In addition to expanding the dataset, we will continue to explore other areas
to apply machine learning to binary object code.  Two interesting areas of 
research include code attribution, and automated reverse engineering techniques
such as determining function boundaries. We feel that machine learning could
play an important role in advancing these research areas. 

\section{Acknowledgments}
The authors would like to thank Dr. Tim Oates of UMBC for guidance on 
machine learning techniques, and Brad Barrett and Charles Lepple of
JHU/APL for discussion and insight into previous research in this area.
Additionally, we would like to thank the reviewers for their comments
and help preparing this paper for publication.

\begin{table*}[ht!]
	\centering
	\begin{tabular}{ l  l  p{9cm} }
		\hline
		Trained Model & WEKA Name & Parameters \\
		\hline
		1-NN & IBk & \texttt{-K 1 -W 0 -A "weka.core.neighboursearch.\-LinearNNSearch -A "weka.core.\-EuclideanDistance -R first-last""}  \\
		\hline
		3-NN & IBk & \texttt{-K {\bf 3} -W 0 -A "weka.core.neighboursearch.\-LinearNNSearch -A "weka.core.\-EuclideanDistance -R first-last""} \\
		\hline
		Decision Tree & J48 & \texttt{-C 0.25 -M 2} \\
		\hline
		Random Tree & RandomTree & \texttt{-K 0 -M 1.0 -V 0.001 -S 1} \\
		\hline
		Random Forest & RandomForest & \texttt{-I 100 -K 0 -S 1 -num-slots 1} \\
		\hline
		Naive Bayes & NaiveBayes & N/A \\
		\hline
		BayesNet & BayesNet & \texttt{-D -Q weka.classifiers.bayes.net.\-search.local.K2 -- -P 1 -S BAYES -E weka.classifiers.bayes.net.\-estimate.SimpleEstimator -- -A 0.5} \\
		\hline
		SVM (SMO) & SMO & \texttt{-C 1.0 -L 0.001 -P 1.0E-12 -N 0 -V -1 -W 1 -K "weka.classifiers.functions.\-supportVector.PolyKernel -E 1.0 -C 250007"} \\
		\hline
		Logistic Regression & SimpleLogistic & \texttt{-I 0 -M 500 -H 50 -W 0.0} \\
		\hline
		Neural Net & MultilayerPerceptron & \texttt{-L 0.3 -M 0.2 {\bf -N 100} -V 0 -S 0 -E 20 {\bf -H 66}} \\
		\hline
	\end{tabular}
	\caption{Full parameter list used for training each WEKA model. 
		Deviations from the default values are marked in bold.}
	\label{table:params}
\end{table*}

\bibliography{paper}
\bibliographystyle{plainnat}
\end{document}